\documentclass[conference]{IEEEtran}
\IEEEoverridecommandlockouts

\usepackage{cite}
\usepackage{amsmath,amssymb,amsfonts}
\usepackage{algorithmic}
\usepackage{graphicx}
\usepackage{textcomp}
\usepackage{xcolor}
\usepackage{caption}
\usepackage{subcaption}
\usepackage{hyperref}
\usepackage{balance}
\usepackage{dblfloatfix}
\usepackage{placeins}

\def\BibTeX{{\rm B\kern-.05em{\sc i\kern-.025em b}\kern-.08em
    T\kern-.1667em\lower.7ex\hbox{E}\kern-.125emX}}

\makeatletter

\def\ps@IEEEtitlepagestyle{
  \def\@oddfoot{\mycopyrightnotice}
  \def\@evenfoot{}
}

\def\mycopyrightnotice{
  {\footnotesize\hfill}
  \gdef\mycopyrightnotice{}
}

\newcommand*\titleheader[1]{\gdef\@titleheader{#1}}

\AtBeginDocument{%
  \let\st@red@title\@title
  \def\@title{%
    \bgroup
    \normalfont\large\centering
    \@titleheader\par
    \egroup
    \vskip1.5em
    \st@red@title
  }
}

\makeatother

\makeatletter

\let\old@ps@headings\ps@headings
\let\old@ps@IEEEtitlepagestyle\ps@IEEEtitlepagestyle

\def\confheader#1{%
  \def\ps@headings{%
    \old@ps@headings
    \def\@oddhead{\strut\hfill#1\hfill\strut}%
    \def\@evenhead{\strut\hfill#1\hfill\strut}%
  }%
  \def\ps@IEEEtitlepagestyle{%
    \old@ps@IEEEtitlepagestyle
    \def\@oddhead{\strut\hfill#1\hfill\strut}%
    \def\@evenhead{\strut\hfill#1\hfill\strut}%
  }%
  \ps@headings
}

\makeatother

\begin{document}

\title{ConformalShift: Targeted Event Reordering Against Adaptive ECG Monitoring}


\author{
Arash Vashagh$^*$ and Yasmin Vashagh$^\dagger$\\
$^*$ Faculty of Computer Science, University of New Brunswick,
Fredericton, New Brunswick E3B 5A3, Canada\\
$^\dagger$ Farzanegan Amin 2 High School, Isfahan, Iran\\
\{arash.vashagh@unb.ca, yasmin.vashagh@gmail.com\}
}

\maketitle

\begin{abstract}
Adaptive conformal prediction can recover clinically important heartbeat
classes missed by a point classifier, but delayed feedback makes its decisions
sensitive to event order. We introduce ConformalShift, a bounded event-reordering
attack that suppresses the ventricular class for rescued events without modifying
ECG waveforms, labels, classifier scores, or the event multiset. ConformalShift
searches for feasible permutations of authentic preceding events that lower the
ventricular threshold before a selected target is evaluated. On disjoint
MIT--BIH confirmation records, the attack suppressed \(66.7\%\) of eligible
targets for Extra Trees and \(60.0\%\) for HistGradientBoosting, compared
with random-schedule rates of \(4.4\%\) and \(12.0\%\), respectively.
Transferred configurations also outperformed random scheduling on INCART,
while reducing the displacement budget weakened the attack on both datasets.
These results show that adaptive monitors in healthcare can be compromised
through the timing of authentic information, even when waveforms, labels,
classifier outputs, and event contents remain unchanged.
\end{abstract}


\section{Introduction}
\label{sec:introduction}

Continuous electrocardiogram monitoring plays an important role in the
early recognition of abnormal cardiac activity. Automated analysis is
especially valuable when clinically significant events are intermittent,
subtle, or embedded within long recordings. Machine learning (ML) has therefore
been increasingly used to support ECG interpretation, including the
detection of atrial fibrillation and other arrhythmias
\cite{10326310,10334666}. In this work, we focus on detecting ventricular ectopic beats, since missing
these abnormal heartbeats may delay clinical review or further assessment. The safety of such systems
depends not only on classifier accuracy, but also on the reliability of the
complete monitoring and decision pipeline.

Adversarial attacks can compromise this reliability. Early studies showed
that carefully crafted input perturbations can cause highly accurate models
to output incorrect predictions
\cite{szegedy2013intriguing,goodfellow2014explaining}. Later work has studied attacks against training data, model privacy,
availability, and interpretability. These attacks may target the data, the model, or the surrounding system
during training or deployment \cite{doi:10.36227/techrxiv.177272853.30003431/v1}. A broad range of
defenses has consequently been proposed to detect adversarial attacks or
increase the robustness of ML models \cite{202607.1022,hasanebrahimi2023density,10326326}. However, most
existing studies assume that the adversary modifies input values, labels,
training samples, model parameters, or internal representations.

Conformal prediction complements a point classifier by returning a set of
plausible labels rather than a single class
\cite{55595013e01f4d0abd12f66422d4289e,pmlr-v235-angelopoulos24a}.
Adaptive conformal inference updates its coverage parameter from observed
errors, allowing it to adapt to changing data streams \cite{gibbs2021adaptive}.
Subsequent work has considered decaying update rates, betting-based
adaptation, multivalid coverage, adversarial sequences, no-regret
connections, and incomplete feedback
\cite{pmlr-v235-angelopoulos24a,pmlr-v235-podkopaev24a,
bastani2022practical,ramalingam2025the,
ge2024stochastic,pmlr-v70-shamir17a}. These studies show that feedback order and availability affect online
coverage, but they do not study targeted reordering under delayed,
class-specific updates.

Existing robust conformal methods primarily address adversarial inputs,
contaminated calibration data, noisy labels, or missing variables
\cite{3692070.3693316,ICLR2024_8759c206,clarkson2024split,
feldman2024robust}. Similarly, prior attacks against ECG classifiers modify
physiological signals to change model decisions
\cite{han2019adversarial,ono2021application,sarkar2024robustness}.
These attacks modify the model inputs, labels, or calibration data. In contrast, the vulnerability studied here preserves
every ECG waveform, verified label, classifier output, and feedback value,
while changing only the order in which authentic events update the adaptive
monitor.

In continuous monitoring, physicians may verify ECG cases after a delay,
and their diagnoses are normally delivered to the adaptive conformal
monitor in first-in, first-out (FIFO) order. We refer to each ECG case together
with its classifier output and verified diagnosis as an \emph{event}.
Because the monitor updates its thresholds using previously received
feedback, its decision for a current case depends not only on the available
information but also on the order in which earlier events were processed.

This order dependence creates a previously understudied attack surface. Existing attacks against ECG systems typically modify physiological signals,
while attacks against conformal prediction commonly perturb inputs, labels,
or calibration data. In a networked monitoring system, however, an attacker
may be able to delay or reorder authentic transmissions without changing
their content. The same ECG signals, classifier outputs, and verified
diagnoses can then produce different conformal decisions. This motivates
studying whether event ordering alone can suppress the monitor's recovery of
a ventricular beat that the point classifier has incorrectly classified as
normal.

\begin{figure*}[t]
    \centering
    \includegraphics[width=0.83\textwidth]{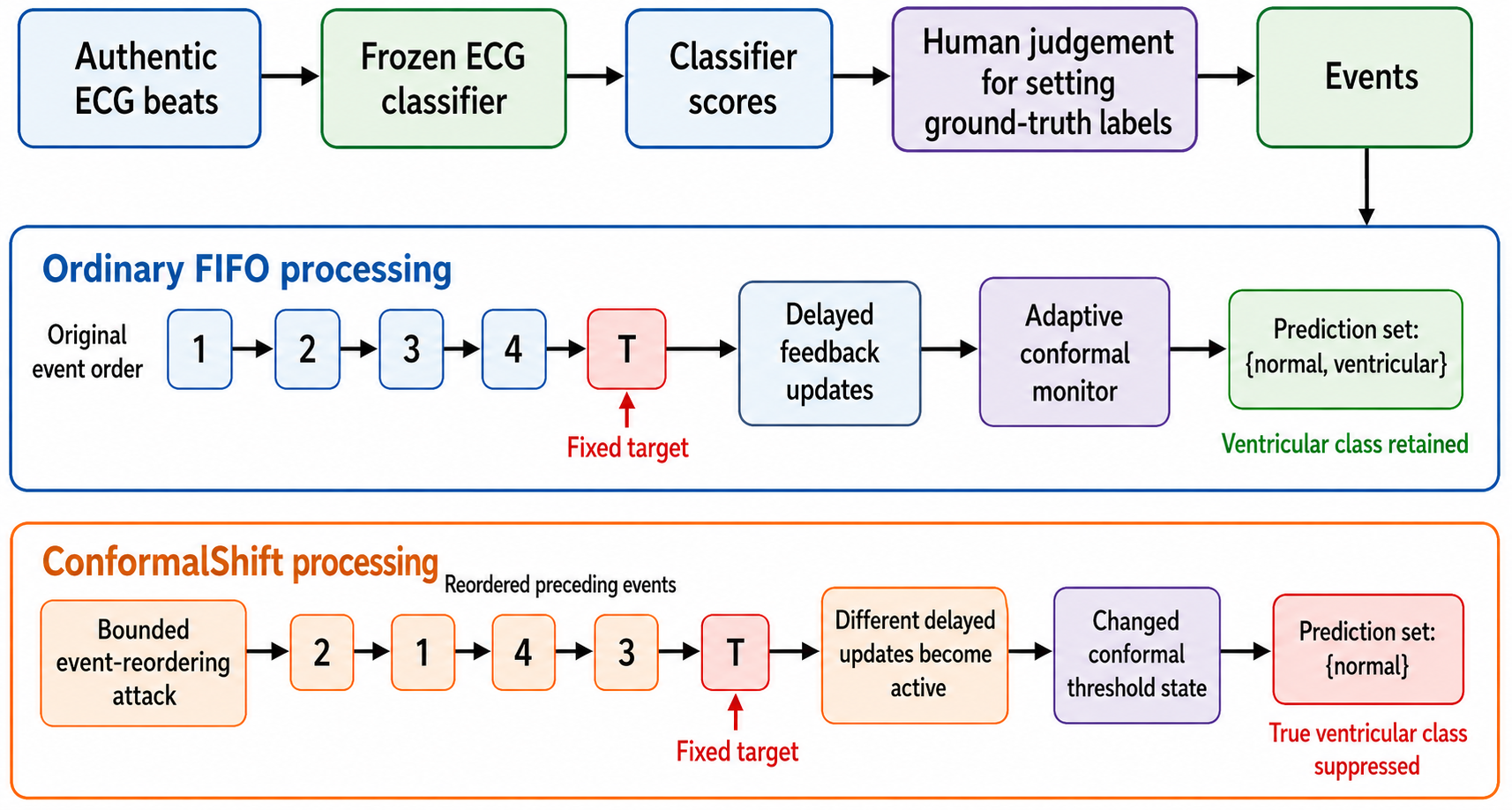}
    \caption{Overview of the ConformalShift threat model. Under ordinary FIFO processing, authentic
    events are handled in their original order and the conformal monitor retains
    the ventricular class for the target beat. ConformalShift instead reorders only the
    preceding authentic events within a bounded displacement budget, changing
    the delayed feedback updates active at the target time and suppressing the
    ventricular class without modifying the ECG signals, labels, classifier
    scores, or target event.}
    \label{fig:ConformalShift_overview}
\end{figure*}

We propose \emph{ConformalShift}, an adversarial sequence manipulation
attack against adaptive conformal ECG monitoring with delayed feedback.
Unlike prior attacks, ConformalShift changes event order to manipulate the
monitor's thresholds without modifying its inputs or classifier. ConformalShift targets ventricular beats that are incorrectly classified as normal
by the point classifier but are recovered by the conformal prediction set
under FIFO processing. It reorders nearby earlier events within a limited range, so different
feedback updates are applied before the target is processed. Figure~\ref{fig:ConformalShift_overview} illustrates the shared ECG processing
pipeline and the difference between ordinary FIFO processing and the ConformalShift
attack.

Our work has three main contributions. First, we formulate bounded event reordering as an attack that changes the
threshold state of an adaptive conformal monitor. Second, we develop a constrained search
method that identifies feasible event orders likely to suppress the target
class. Third, we evaluate ConformalShift across two ECG datasets and two frozen
classifier families, compare it with random feasible schedules, and analyze
how the permitted displacement budget affects attack success.\footnote{Code and reproduction scripts:
\url{https://github.com/arashVsh/ConformalShift-adversarial-ecg}.}

\section{Method}
\label{sec:method}

This section describes the adaptive class-conditional conformal monitor and
the ConformalShift attack. We first define the frozen ECG classifier, the conformal
prediction set, and the delayed state-update mechanism. We then define the reordering threat model, the suppression objective, and
the search for an adversarial schedule.

\subsection{Adaptive Class-Conditional Monitoring}
\label{subsec:ConformalShift_monitor}

Let $\mathcal{Y}=\{1,\ldots,K\}$ be the set of $K$ heartbeat classes. At stream position \(t\), the monitoring system receives an ECG beat
\(x_t\). Let \(y_t\in\mathcal{Y}\) denote its verified label, which becomes
available to the adaptive monitor only after the feedback delay defined
below. A frozen classifier produces the
probability vector
$\mathbf{p}_t=(p_{t,1},\ldots,p_{t,K})$, where
$\sum_{c\in\mathcal{Y}}p_{t,c}=1$. For candidate class
$c\in\mathcal{Y}$, the nonconformity score is $a_{t,c}=1-p_{t,c}$.

The monitor maintains one adaptive threshold for each class. Let
$\mathbf{q}_t=(q_{1,t},\ldots,q_{K,t})$ denote the threshold vector used at
position $t$. The resulting class-conditional prediction set is

\begin{equation}
\mathcal{C}_t
=
\left\{
c\in\mathcal{Y}
\;\middle|\;
a_{t,c}\leq q_{c,t}
\right\}.
\label{eq:ConformalShift_prediction_set}
\end{equation}

Class \(c\) is included when its nonconformity score is at most the current
threshold \(q_{c,t}\). The classifier
is not updated during deployment; only the threshold vector evolves.

Let $\alpha\in(0,1)$ be the desired miscoverage rate, $\eta>0$ the
adaptation rate, $\rho\in(0,1]$ the recency factor, and $q^{(0)}$ the
initial threshold. All class thresholds are initialized as
$q_{c,0}=q^{(0)}$. Before processing feedback at position \(t\), each threshold decays toward
its initial value:

\begin{equation}
\bar{q}_{c,t}
=
q^{(0)}
+
\rho
\left(
q_{c,t-1}-q^{(0)}
\right).
\label{eq:ConformalShift_recency}
\end{equation}

Let $\delta\geq0$ be the fixed feedback delay. Feedback generated by the
event observed at position $i$ becomes available immediately before the
prediction at position $i+\delta+1$. The feedback events that mature at
position $t$ therefore form the set $\mathcal{D}_t
=
\left\{
i<t
\;\middle|\;
i+\delta+1=t
\right\}$.

For each event, define its conformal miss indicator as

\begin{equation}
e_i
=
\mathbb{I}
\left[
a_{i,y_i}>q_{y_i,i}
\right],
\label{eq:ConformalShift_miss_indicator}
\end{equation}

where $\mathbb{I}[\cdot]$ equals one when its argument is true and zero
otherwise. The indicator is fixed when
event $i$ is originally evaluated and is applied only when its verified
feedback becomes available.

After the recency operation in \eqref{eq:ConformalShift_recency}, the threshold for
class $c$ is updated according to

\begin{equation}
\begin{aligned}
q_{c,t}
=
\operatorname{clip}_{[q_{\min},q_{\max}]}
\Bigg(
\bar{q}_{c,t}
+
\eta
\sum_{i\in\mathcal{D}_t}
\mathbb{I}[y_i=c]
\left(
e_i-\alpha
\right)
\Bigg),
\end{aligned}
\label{eq:ConformalShift_threshold_update}
\end{equation}

where $q_{\min}$ and $q_{\max}$ are the lower and upper threshold bounds,
respectively, and $\operatorname{clip}_{[q_{\min},q_{\max}]}(\cdot)$
projects its argument onto that interval. A matured miss increases the
threshold of the corresponding verified class by
$\eta(1-\alpha)$, whereas a matured covered event decreases it by
$\eta\alpha$. Feedback associated with one class does not directly update
the thresholds of the other classes.

\subsection{Target Selection and Threat Model}
\label{subsec:ConformalShift_threat}

ConformalShift targets the ventricular class, denoted by
\(v\in\mathcal{Y}\). Let \(t^\star\) be the stream position of the selected
ventricular event, so that \(y_{t^\star}=v\). The attack considers events
that are missed by the classifier's top-one decision but rescued by the
class-conditional prediction set under the original FIFO order, satisfying
\(\arg\max_{c\in\mathcal{Y}}p_{t^\star,c}\neq v\) and
\(v\in\mathcal{C}_{t^\star}^{\mathrm{fifo}}\). The superscript
\(\mathrm{fifo}\) denotes the monitor trajectory obtained from the original
event order. These are cases where the conformal set recovers a ventricular class missed
by the point classifier.

The attacker can modify the transmission order of the $W$ authentic events
immediately preceding the target. The target event itself remains fixed.
The original local positions of the preceding events are
$1,\ldots,W$. A candidate transmission schedule is represented by the
permutation
$\boldsymbol{\pi}=(\pi_1,\ldots,\pi_W)$, where $\pi_j$ is the original
local index of the event transmitted at attacked position $j$.

Let $\mathfrak{S}_W$ denote the set of all permutations of
$\{1,\ldots,W\}$, and let $d\geq0$ be the maximum permitted displacement.
The feasible schedule set is

\begin{equation}
\begin{aligned}
\Pi_d
=
\Bigg\{
\boldsymbol{\pi}\in\mathfrak{S}_W
\;\Bigg|\;
\max_{1\leq j\leq W}
\left|
\pi_j-j
\right|
\leq d
\Bigg\}.
\end{aligned}
\label{eq:ConformalShift_feasible_schedules}
\end{equation}

Constraint \eqref{eq:ConformalShift_feasible_schedules} limits each event's displacement to \(d\) positions. Since
$\boldsymbol{\pi}$ is a permutation, the attacker cannot create, delete,
duplicate, or modify an event. In particular, the ECG waveforms, verified
labels, classifier probabilities, and event multiset remain unchanged.
Only the temporal order of the preceding events is altered.

For each candidate schedule \(\boldsymbol{\pi}\), we replay the monitor
using the update rules in
\eqref{eq:ConformalShift_recency}--\eqref{eq:ConformalShift_threshold_update}. Let
$\mathbf{q}_{t^\star}^{\boldsymbol{\pi}}$ be the threshold vector available
when the unchanged target event is evaluated after this replay.

\subsection{Targeted Suppression Objective}
\label{subsec:ConformalShift_objective}

Because the target waveform and classifier are unchanged, its
nonconformity score $a_{t^\star,v}$ is invariant across all feasible
schedules. Because reordering changes only the ventricular threshold, we define the
suppression margin as

\begin{equation}
m(\boldsymbol{\pi})
=
a_{t^\star,v}
-
q_{v,t^\star}^{\boldsymbol{\pi}}.
\label{eq:ConformalShift_suppression_margin}
\end{equation}

The ventricular class is included in the target prediction set when
$m(\boldsymbol{\pi})\leq0$ and excluded when
$m(\boldsymbol{\pi})>0$. Due to the target-selection condition, its FIFO margin is nonpositive.
Hence, a successful ConformalShift attack must satisfy

\begin{equation}
\begin{aligned}
m(\boldsymbol{\pi}^{\mathrm{fifo}})
&\leq 0, m(\boldsymbol{\pi})
&>0,
\end{aligned}
\label{eq:ConformalShift_success_condition}
\end{equation}

where
$\boldsymbol{\pi}^{\mathrm{fifo}}=(1,\ldots,W)$ is the original local
schedule.

The attack searches for the feasible schedule with the largest suppression
margin  $\boldsymbol{\pi}^{\star}
=
\arg\max_{\boldsymbol{\pi}\in\Pi_d}
m(\boldsymbol{\pi})$.

This encourages schedules that reduce
$q_{v,t^\star}^{\boldsymbol{\pi}}$ below the fixed target score. The attack
can achieve this by changing which class-$v$ feedback events mature before
the target. In particular, covered ventricular events decrease the
ventricular threshold when their feedback matures, whereas ventricular
misses increase it. The attack can move covered ventricular feedback earlier or ventricular
misses later, without changing either event.

For each target, the threshold displacement induced by the selected
schedule is measured as $\Delta q_{t^\star}
=
q_{v,t^\star}^{\boldsymbol{\pi}^{\star}}
-
q_{v,t^\star}^{\mathrm{fifo}}$.

A negative value of $\Delta q_{t^\star}$ indicates that the attack lowers
the ventricular threshold. Since the target score remains fixed, the
corresponding gain in suppression margin is

\begin{equation}
\Delta m_{t^\star}
=
m(\boldsymbol{\pi}^{\star})
-
m(\boldsymbol{\pi}^{\mathrm{fifo}})
=
-\Delta q_{t^\star}.
\label{eq:ConformalShift_margin_gain}
\end{equation}

\subsection{Constrained Schedule Search}
\label{subsec:ConformalShift_search}

Because there are \(W!\) possible orders, exhaustive search is impractical. ConformalShift therefore
uses a displacement-aware beam search followed by local refinement. At
attacked position $j$, a remaining event with original local index $i$ can
be appended only when $|i-j|\leq d$. An event is scheduled immediately if delaying it would violate the
displacement limit.

Each partial schedule is evaluated by replaying the monitor state up to the
corresponding attacked position. Beam candidates are ranked according to
their estimated effect on the ventricular threshold at the target. We retain candidates with different ventricular thresholds and pending
ventricular feedback states to avoid duplicate search paths.

After the beam search produces complete feasible schedules, the candidate
pool is enlarged using the FIFO order, target-aware heuristic schedules,
and randomly generated feasible permutations. Each candidate is then
refined through legal schedule modifications, including adjacent swaps,
long-range swaps, and event relocations. A proposed modification is accepted
only when it remains in $\Pi_d$ and increases the exact margin in
\eqref{eq:ConformalShift_suppression_margin}. We use multiple random restarts to explore different initial schedules.

\begin{figure*}[!t]
    \centering
    \includegraphics[width=0.97\textwidth]{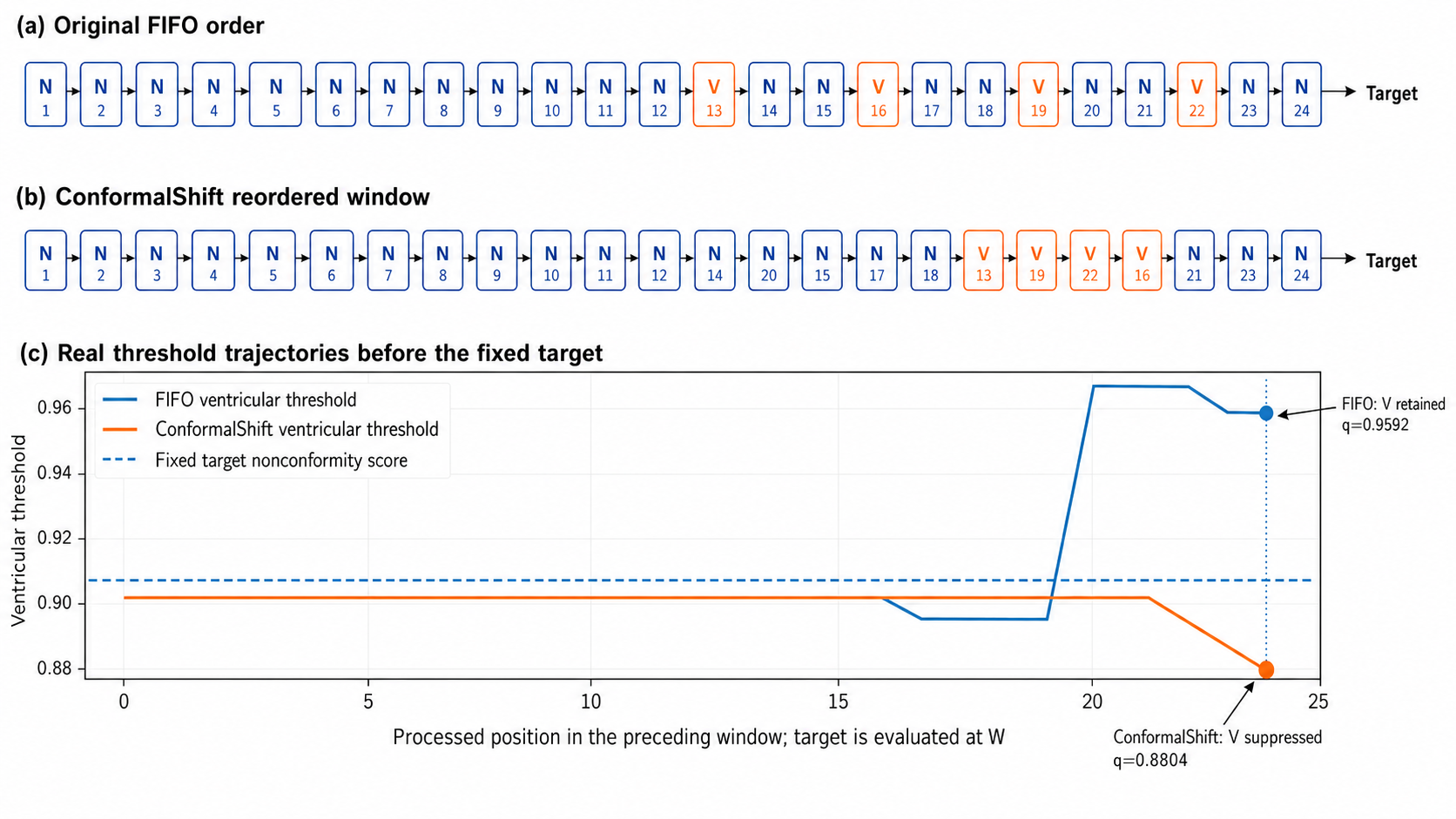}
    \caption{Successful ConformalShift attack on MIT--BIH record 214. Panels (a)
    and (b) show the original FIFO order and the adversarially reordered
    preceding-event window, respectively. Each box reports the heartbeat
    class and its original local position, while the target event remains
    fixed. Panel (c) shows the corresponding ventricular-threshold
    trajectories. The target nonconformity score is unchanged, but ConformalShift
    lowers the ventricular threshold from \(0.9592\) to \(0.8804\), changing
    the target margin from \(-0.0528\) to \(0.0260\) and excluding the
    ventricular class from the prediction set.}
    \label{fig:ConformalShift_success_case}
\end{figure*}

The final adversarial schedule is the feasible candidate with the largest
exact suppression margin. A target is counted as successfully suppressed
only when the selected schedule satisfies
\eqref{eq:ConformalShift_success_condition}; improvements that leave the
ventricular class inside the prediction set are not counted as successful
attacks. Figure~\ref{fig:ConformalShift_success_case} illustrates a real successful attack.

\begin{table*}[t]
\centering
\caption{Clean FIFO performance averaged over five random seeds.
V F1 and V coverage refer to the ventricular class. V Rescue is the
fraction of point-classifier ventricular misses recovered by the conformal
monitor.}
\label{tab:clean_utility}
\begin{tabular}{llcccccc}
\hline
Dataset & Victim & Accuracy & V F1 & Coverage &
V Coverage & Set Size & V Rescue \\
\hline
MIT--BIH & Extra Trees
& 0.9329 & 0.8229 & 0.9419 & 0.9691 & 1.5454 & 0.6995 \\

MIT--BIH & HistGradientBoosting
& 0.9200 & 0.7287 & 0.9463 & 0.9245 & 1.2237 & 0.7316 \\

INCART & Extra Trees
& 0.9675 & 0.8957 & 0.9569 & 0.9727 & 1.3598 & 0.7706 \\

INCART & HistGradientBoosting
& 0.9194 & 0.8539 & 0.9446 & 0.9449 & 1.1459 & 0.6274 \\
\hline
\end{tabular}
\end{table*}

\section{Results}
\label{sec:results}

We evaluate ConformalShift using clean utility, disjoint MIT--BIH
confirmation records, descriptive full-DS2 results, INCART transfer, and a
displacement-budget ablation.

\begin{table*}[t]
\centering
\caption{ConformalShift results. MIT--BIH confirmation uses disjoint
records; full DS2 is descriptive; INCART uses transferred configurations.
Random is the mean over 128 feasible schedules per target.}
\label{tab:main_attack_results}
\begin{tabular}{lllcccccc}
\hline
Dataset & Evaluation & Victim & \(W\) & \(d\) & \(\delta\) &
Successes & Targeted & Random \\
\hline
MIT--BIH & Confirmation & Extra Trees
& 24 & 8 & 4 & 4/6 & 66.7\% & 4.4\% \\

MIT--BIH & Confirmation & HistGradientBoosting
& 16 & 4 & 2 & 6/10 & 60.0\% & 12.0\% \\

MIT--BIH & Full DS2 & Extra Trees
& 24 & 8 & 4 & 17/27 & 63.0\% & 7.7\% \\

MIT--BIH & Full DS2 & HistGradientBoosting
& 16 & 4 & 2 & 35/71 & 49.3\% & 10.8\% \\

INCART & Transfer & Extra Trees
& 24 & 8 & 4 & 5/15 & 33.3\% & 5.0\% \\

INCART & Transfer & HistGradientBoosting
& 16 & 4 & 2 & 5/15 & 33.3\% & 4.5\% \\
\hline
\end{tabular}
\end{table*}

\subsection{Experimental Setup}
\label{subsec:experimental_setup}

We evaluated ConformalShift on the MIT--BIH Arrhythmia Database and the
St.\ Petersburg INCART 12-lead Arrhythmia Database. Beat annotations were
mapped to the five AAMI classes
\(\{\mathrm{N},\mathrm{S},\mathrm{V},\mathrm{F},\mathrm{Q}\}\).
For MIT--BIH, we used the first ECG channel and extracted \(360\)-sample
windows centred on each annotated beat. For INCART, we used lead II when
available, extracted one-second windows centred on the annotations, and
resampled them to \(360\) samples. Each window was normalized by subtracting
its median and dividing by its standard deviation, then represented by
\(21\) amplitude, derivative, quantile, peak-location, and spectral
features. MIT--BIH used the standard inter-patient DS1--DS2 split. INCART
used a fixed patient-independent split of 22 training and 10 test patients,
generated with seed 2026.

We trained frozen Extra Trees and HistGradientBoosting classifiers. The
conformal monitor used \(\alpha=0.1\), \(\eta=0.08\),
\(q^{(0)}=0.92\), and \(\rho=0.995\). The locked Extra Trees configuration
used \(W=24\), \(d=8\), and \(\delta=4\), while HistGradientBoosting used
\(W=16\), \(d=4\), and \(\delta=2\). Configurations were selected on
discovery records, frozen before evaluation on disjoint MIT--BIH
confirmation records, and transferred to INCART without dataset-specific
attack tuning. The search used a beam width of \(256\), \(10\) local-search
passes, \(24\) random restarts, \(256\) neighborhood candidates, and at most
three candidates per state signature; simulated annealing was disabled.
For each target, the random baseline evaluated \(128\) feasible schedules
under the same \(W\), \(d\), and \(\delta\) constraints.

INCART transfer used \(15\) selected targets per victim, model seed \(7\),
and attack seed \(1007\); the reported attack rates are single-seed results.

\subsection{Clean Monitoring Utility}
\label{subsec:clean_utility}

Before evaluating the attack, we measured the frozen classifiers and the
adaptive conformal monitor under the original FIFO sequence.
Table~\ref{tab:clean_utility} reports classifier accuracy, ventricular-class
F1-score, overall conformal coverage, ventricular coverage, average
prediction-set size, and ventricular rescue rate. Rescue is the fraction of
true ventricular events missed by the point classifier but retained by the
conformal prediction set.

The conformal monitor recovered many ventricular events missed by the point
classifiers. On MIT--BIH, it rescued \(70.0\%\) and \(73.2\%\) of
ventricular misses for Extra Trees and HistGradientBoosting, respectively,
while ventricular coverage remained above \(92\%\). On INCART, it rescued
\(77.1\%\) of Extra Trees misses and \(62.7\%\) of HistGradientBoosting
misses, with average prediction-set sizes of \(1.36\) and \(1.15\),
respectively.

\subsection{Attack Results}
\label{subsec:attack_results}

Table~\ref{tab:main_attack_results} summarizes ConformalShift across three
evaluation settings: disjoint MIT--BIH confirmation records, the descriptive
full-DS2 evaluation, and cross-dataset transfer to INCART. The confirmation
records were not used during attack-configuration discovery. The full-DS2
results are descriptive because they include discovery records, while the
INCART results use the locked MIT--BIH configurations without
dataset-specific attack tuning.

On the disjoint MIT--BIH confirmation records, ConformalShift suppressed
four of six eligible Extra Trees targets, corresponding to a targeted rate
of \(66.7\%\). Random feasible schedules suppressed only \(4.4\%\) of
targets on average, producing a gap of \(62.2\) percentage points. The
\(95\%\) Wilson interval for the targeted rate was \(30.0\%-90.3\%\).
The mean suppression-margin gain was \(0.0491\), and successful attacks
ended with a mean positive margin of \(0.0283\).

Against HistGradientBoosting, ConformalShift suppressed six of ten
confirmation targets, compared with a random-schedule rate of \(12.0\%\).
The targeted-over-random gap was \(48.0\) percentage points, and the
\(95\%\) Wilson interval was \(31.3\%-83.2\%\). The mean margin gain was
\(0.0438\), while the mean final margin among successful attacks was
\(0.0302\). These results show that attack success depends on selecting
specific event orders rather than merely sampling among feasible schedules.

Across all 22 MIT--BIH DS2 records, ConformalShift suppressed \(17/27\)
Extra Trees targets and \(35/71\) HistGradientBoosting targets. These results
are descriptive rather than independent confirmation, but the targeted rates
remained substantially above the corresponding random baselines.

The locked MIT--BIH configurations also transferred to INCART. Both victim
models suppressed \(5/15\) selected targets, corresponding to a targeted
rate of \(33.3\%\), compared with random-schedule rates of \(5.0\%\) for
Extra Trees and \(4.5\%\) for HistGradientBoosting. The mean margin gains
were \(0.0141\) and \(0.0260\), respectively, and successful attacks ended
with positive mean margins. These results provide evidence of cross-dataset
transfer without INCART-specific attack tuning, although the \(15\)-target,
single-seed evaluation remains uncertain. Because the two victims used
different locked \(W\), \(d\), and \(\delta\) values, their results should
not be interpreted as a controlled comparison of intrinsic model
vulnerability.

\subsection{Displacement-Budget Ablation}
\label{subsec:ablation}

We evaluated how the maximum displacement budget affects ConformalShift
using the Extra Trees victim. For MIT--BIH, the comparison used the same
six confirmation targets, window size, delay, model predictions, and search
configuration; only \(d\) was changed. The same locked settings were used
for INCART. Table~\ref{tab:displacement_ablation} compares the original and
reduced displacement budgets on both datasets.

\begin{table}[t]
\centering
\caption{Effect of the maximum displacement budget for Extra Trees.
Margin gain is the mean increase in the target suppression margin.}
\label{tab:displacement_ablation}
\begin{tabular}{lccccc}
\hline
Dataset & \(d\) & Successes & Targeted & Random & Margin Gain \\
\hline
MIT--BIH & 8 & 4/6  & 66.7\% & 4.4\% & 0.0491 \\
MIT--BIH & 4 & 1/6  & 16.7\% & 1.4\% & 0.0144 \\
INCART   & 8 & 5/15 & 33.3\% & 5.0\% & 0.0141 \\
INCART   & 4 & 4/15 & 26.7\% & 4.5\% & 0.0085 \\
\hline
\end{tabular}
\end{table}

Reducing the displacement budget weakened ConformalShift on both datasets.
On the paired MIT--BIH targets, targeted suppression decreased from
\(66.7\%\) to \(16.7\%\), while the mean margin gain fell from \(0.0491\)
to \(0.0144\). On INCART, targeted suppression decreased from \(33.3\%\)
to \(26.7\%\), and the mean margin gain declined from \(0.0141\) to
\(0.0085\). Therefore, a larger displacement budget gives the attack more
freedom to move covered ventricular feedback earlier and ventricular misses
later.

\section{Discussion}
\label{sec:discussion}

The results reveal a temporal-integrity vulnerability in adaptive conformal
monitoring. Protecting model
inputs and parameters is insufficient when downstream decisions depend on
the order in which authentic feedback is incorporated.

The large targeted--random gap shows that the vulnerability is structured,
not a generic consequence of reordering. ConformalShift performs adversarial state steering by selecting orders
that change which ventricular updates mature before the target while
preserving all event contents.

The clean results also distinguish aggregate coverage from event-level
safety: despite high coverage and frequent ventricular rescue, selected
rescued events can still be suppressed. Adaptive conformal systems should
therefore undergo targeted, class-specific, and sequence-aware stress tests.

The transferred configurations remained stronger than random scheduling on
INCART, suggesting that the vulnerability is not specific to one dataset,
although its strength depends on the local event composition and threshold
trajectory. The displacement ablation further shows that restricting
out-of-order delivery can reduce attack effectiveness. Practical defenses
could authenticate timestamps, reject
stale feedback, or limit unusually large threshold changes.

This study is limited to a white-box attacker, retrospective ECG datasets,
one delayed class-conditional update rule, and relatively small confirmatory
target sets. Future work should examine partial-knowledge attacks, uncertain
feedback delays, other adaptive conformal methods, additional target
classes, and defenses designed explicitly for temporal integrity.

\section{Conclusion}
\label{sec:conclusion}

We introduced ConformalShift, a bounded event-reordering attack against
adaptive conformal ECG monitoring with delayed feedback. The attack
suppresses ventricular rescue events by reordering only authentic preceding
events, while preserving the ECG waveforms, verified labels, classifier
outputs, event multiset, and target event. Its advantage over
random feasible schedules, transfer from MIT--BIH to INCART, and degradation
under tighter displacement limits show that adaptive systems can be attacked by
changing when valid information is processed. Therefore, robustness evaluations for adaptive healthcare monitoring
should extend beyond attacks on inputs and models to include
feedback order, delay, and temporal integrity. Defenses that authenticate
timestamps, constrain out-of-order delivery, and monitor abnormal state
transitions may be necessary to protect the safety benefits provided by
adaptive conformal prediction.

\bibliographystyle{IEEEtran}
\bibliography{IEEEabrv,references}

\end{document}